\documentclass[journal]{IEEEtran}
\usepackage{graphicx}
\usepackage{algorithmic}

\usepackage{amssymb}
\usepackage{amsmath,amsfonts}
\usepackage{multirow}
\usepackage{cite}
\usepackage{bm}
\usepackage{booktabs}
\usepackage{caption}
\usepackage{color}
\usepackage{subfigure}
\usepackage{subcaption}
\usepackage{comment}

\usepackage{textcomp}                                         
\usepackage{float} 
\usepackage{bbm}
\usepackage{hyperref}

\author{Zhihao Chen\inst{1} \and
Yang Zhou\inst{2} \and
Anh Tran\inst{3} \and
Junting Zhao\inst{1} \and
Liang Wan\inst{1} \and
Gideon Ooi\inst{3} \and
Lionel Cheng\inst{3} \and
Choon Hua Thng\inst{3} \and
Xinxing Xu\inst{2} \and 
Yong Liu \inst{2} \and
Huazhu Fu\inst{2}}
\begin{document}
\title{Topicwise Separable Sentence Retrieval for Medical Report Generation} 
\author{Junting Zhao, Yang Zhou, Zhihao Chen, Huazhu Fu, Liang Wan$^{\dag}$
\thanks{L. Wan is the corresponding author.}
\thanks{J. Zhao, Z. Chen, and L. Wan are with the College of Intelligence and Computing, Tianjin University, Tianjin 300350, China. (e-mail: zhaojt@tju.edu.cn, zh\_chen@tju.edu.cn, lwan@tju.edu.cn)}
\thanks{Y. Zhou and H. Fu are with the Institute of High Performance Computing (IHPC), Agency for Science, Technology and Research (A*STAR), Singapore 138632. (e-mail: zhou\_yang@ihpc.a-star.edu.sg, hzfu@ieee.org)}
}

\maketitle

\begin{abstract}
Automated radiology reporting holds immense clinical potential in alleviating the burdensome workload of radiologists and mitigating diagnostic bias.
Recently, retrieval-based report generation methods have garnered increasing attention due to their inherent advantages in terms of the quality and consistency of generated reports.
However, due to the long-tail distribution of the training data, these models tend to learn frequently occurring sentences and topics, overlooking the rare topics.
Regrettably, in many cases, the descriptions of rare topics often indicate critical findings that should be mentioned in the report.
To address this problem, we introduce a Topicwise Separable Sentence Retrieval (Teaser) for medical report generation.
To ensure comprehensive learning of both common and rare topics, we categorize queries into common and rare types to learn differentiated topics, and then propose Topic Contrastive Loss to effectively align topics and queries in the latent space.
Moreover, we integrate an Abstractor module following the extraction of visual features, which aids the topic decoder in gaining a deeper understanding of the visual observational intent.
Experiments on the MIMIC-CXR and IU X-ray datasets demonstrate that Teaser surpasses state-of-the-art models, while also validating its capability to effectively represent rare topics and establish more dependable correspondences between queries and topics.
\end{abstract}


\IEEEpeerreviewmaketitle

\section{Introduction}
\label{sec:introduction}
Radiology medical reports provide detailed descriptions of patients' conditions and diagnostic results, playing a key role in medical diagnosis and treatment. However, writing medical reports requires radiologists to possess extensive medical knowledge and a specialized background. Due to the large amount of medical reports, the task of writing them is highly laborious and time-consuming. Moreover, variations in knowledge and experience among radiologists can lead to biases and errors in the reports. Therefore, the automation of generating high-quality reports is of crucial importance.

\begin{figure}[!t]
\centering
\includegraphics[width=1\linewidth]{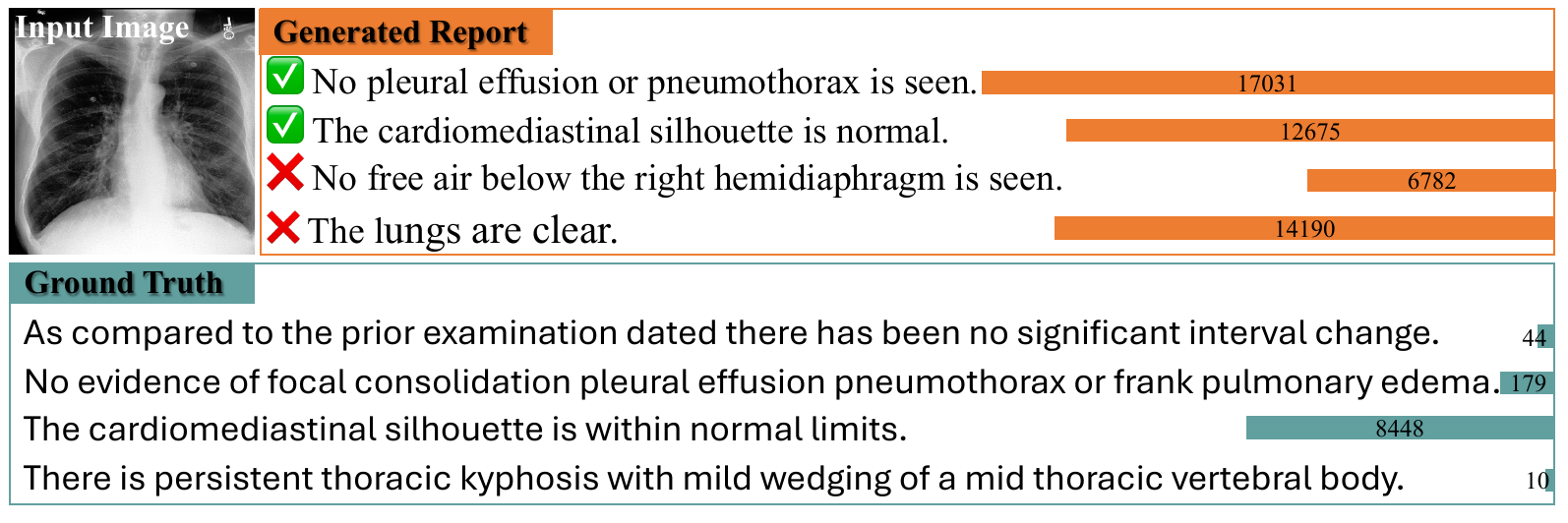}
\caption{
Illustration of the common mistakes made by existing retrieval-based methods.
The histogram represents the sentence frequencies in the training set. Existing methods tend to retrieve high-frequency sentences while neglecting low-frequency ones.}
\label{Fig:intro_case}
\vspace{-6mm}
\end{figure}

The task of medical report generation originated from image captioning~\cite{vinyals2015show, krause2017hierarchical,li2019entangled}, placing greater emphasis on longer and more accurate contextual descriptions. This poses heightened requirements for both the length and precision of generated reports. Current medical report generation approaches can be grouped into two main categories: generation-based and retrieval-based methods.
Generation-based methods consider text generation as a sequential modeling problem, which aims to predict the next word in a sequence given the preceding textual and visual context (radiology image).
Previous generation-based works adapt RNN~\cite{jing2018automatic, li2018hybrid} or Transformers~\cite{R2Gen,zhang2020radiology,MedXChat2023arXiv} for report generation. 
Some works introduced memory mechanisms~\cite{R2Gen, multicriteria2022TMI} and knowledge graphs~\cite{zhang2020radiology,KGAE,liu2021exploring} to make the text generation align with clinical writing styles and prior medical knowledge. 
%
Moreover, recent researches~\cite{MedXChat2023arXiv,RaDialog2023arXiv} are exploring the potential of replacing the decoder with pretrained LLMs (Large Language Models) that have been trained on vast amounts of medical literature and knowledge databases. 
However, generation-based approaches face two notable limitations. Firstly, they often focus more on generating coherent clinical descriptions, potentially at the expense of accurately identifying abnormalities in accompanying images.
Secondly, the inherent freedom in organizing language can lead to illusionary issues in generated content, where the generated information may exhibit logical inaccuracies or deviate from the actual circumstances.

On the other hand, retrieval-based report generation methods~\cite{endo2021retrieval, han2018towards} perform text generation by searching through the sentence gallery to find sentences or passages that are relevant to the input query.
The retrieved information is then utilized to construct a coherent medical report. 
In contrast to generation-based methods, retrieval-based methods do not need to train the model from scratch to generate coherent sentences, and focus on matching medical findings in the image with the corresponding textual descriptions.
For example, TranSQ~\cite{transq} utilizes the Transformer encoder-decoder architecture to learn topic-level image representations as the queries. It then retrieves relevant sentences from a gallery containing embedded sentences in the training set to construct the final report.
Despite its simplicity, TranSQ outperforms the majority of generation-based methods in both natural language generation (NLG) and clinical efficacy (CE) metrics, which demonstrates the great potential of retrieval-based approaches in the medical report generation task.

However, upon closer examination of the reports generated by existing retrieval-based methods, we observed that they tends to produce high-frequency sentences in the training set and overlook low-frequency sentences (see Figure~\ref{Fig:intro_case}).
This observation could be attributed to the long-tail distribution present in the data, which makes these methods prefer generating sentences that appear more frequently in the training set while overlooking a comprehensive understanding of the rare topics.
In many cases, the descriptions of rare topics often indicate critical findings that should be mentioned in the generated report.
Additionally, existing methods fail to learn the relationships between queries and topics, where different queries may map to the same topic, resulting in repetitive sentences with identical semantics appearing in the reports.

To address the problem of retrieval-based report generation, this paper proposes a \textbf{T}opicwise S\textbf{e}p\textbf{a}rable \textbf{Se}ntence \textbf{R}etrieval (\textbf{Teaser}) method for medical report generation. In order to facilitate the learning of both high and low-frequency topics, we categorize queries into two groups: common queries and rare queries. Common queries are responsible for retrieving general and widely applicable topics, such as general medical terminology and descriptions of common diseases. On the other hand, rare queries are designed to retrieve specific topics, mainly focusing on descriptions of uncommon diseases. During the training process, we categorize common and rare queries and align them with their respective sentence embeddings, effectively mitigating the influence of the long-tail distribution in the dataset. To establish a more explicit correspondence between queries and topics, we propose a Topic Contrastive Loss (TCL), which brings semantically similar queries closer together in the latent space while pushing dissimilar queries farther apart. 
Furthermore, we adopt the Abstractor module to compress visual features and reduce noise in visual embeddings, which facilitates better alignment and matching between visual and textual findings.


In summary, our main contributions are as follows:
\begin{enumerate}
    \item We introduce a Topicwise Separable sentence retrieval method, which tackles the challenge of incomplete predictions arising from the long-tail distribution of data by separating the retrieval of common and rare topics.
    \item We propose the Topic Contrastive Loss, which explicitly aligns queries and topics with similar semantics, thereby mitigating the many-to-one and one-to-many issues.
    \item We adopt the Abstractor, which aids the topic decoder in achieving a deeper understanding of visual observation intentions by compressing visual features and eliminating visual noise.
    \item Our proposed method achieves state-of-the-art performance for both NLG and CE metrics on two well-known medical report generation benchmarks.
\end{enumerate}

\section{Related Works}
\subsection{Generation-based Medical Report Generation}

The task of radiology report generation has its roots in image captioning~\cite{you2016image, liang2017recurrent}, prompting most early efforts to employ the CNN-RNN architecture. Unlike typical image captioning tasks, radiology reports tend to be lengthier and comprise multiple sentences, each addressing distinct abnormal findings. Consequently, some researchers turned to a Hierarchical Recurrent Network (HRNN)~\cite{jing2018automatic, li2018hybrid} to facilitate state transitions within a hierarchical framework, addressing the challenge of handling long-text dependencies. This HRNN framework encompasses both a topic RNN and a sentence RNN. The former is responsible for generating topic vectors, which are subsequently utilized by the sentence RNN to craft descriptive sentences for the images. Nevertheless, these approaches often necessitate manual template extraction or prior medical knowledge to guide the state transition process. This not only incurs significant human efforts but also constrains the model's ability to generalize effectively.

\begin{figure*}[!t]
\centering
\includegraphics[width =1\linewidth]{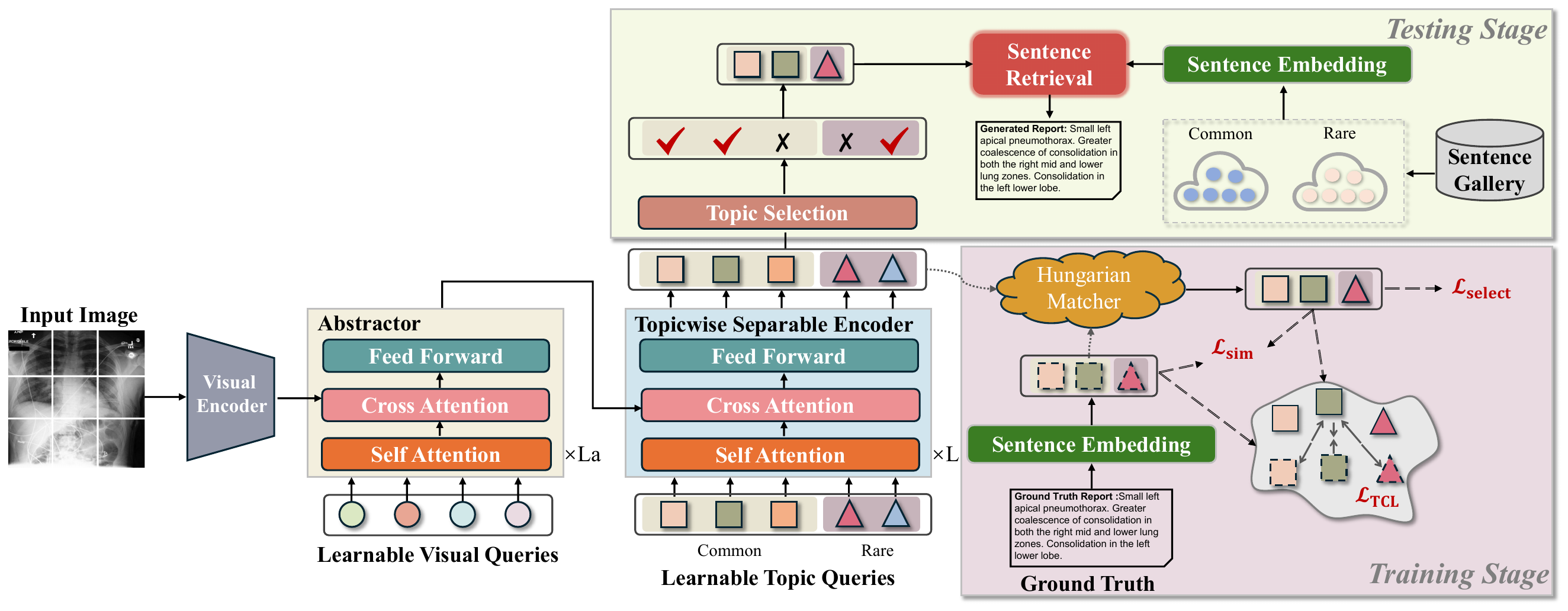}
\caption{
The framework of the proposed Teaser includes the visual encoder and Abstractor, responsible for extracting and abstracting visual features, and the Topicwise Separable Encoder for obtaining topic embeddings. In the training stage, Hungarian Matcher is utilized to select the best matching topic embedding for each ground truth sentence. The similarity loss ($\mathcal{L}_{\mathrm{sim}}$) and topic contrastive loss ($\mathcal{L}_{\mathrm{TCL}}$) are used to align these embeddings, while the selection($\mathcal{L}_{\mathrm{select}}$) loss aids in topic filtering. In the testing stage, the chosen topic embeddings retrieve the best matches separately from the common and rare sentence galleries. These matches are then merged together to generate the final report.}
\label{Fig:framework}
\vspace{-4.1mm}
\end{figure*}

Recently, the Transformer architecture~\cite{vaswani2017attention} has gained significant traction in report generation tasks~\cite{KnowMT2022MIA, li2022self, li2022cross} owing to its remarkable parallel processing capabilities and proficiency in capturing long-range dependencies. To emulate the reporting habits of medical professionals, some researchers have introduced a series of memory mechanisms. Chen~\textit{et al.}~\cite{chen2020r2gen} devised a memory-driven conditional layer normalization (MCLN) to integrate relational memory into the Transformer, with a focus on templating and accommodating long-form text characteristics. Similarly, Chen~\textit{et al.}~\cite{chen2021r2gencmn} employed a memory matrix to store cross-modal information, utilizing it for memory querying and responding for both visual and textual features. Additionally, certain efforts have leveraged existing medical knowledge as a priori information, integrating it in the form of a knowledge graph embedded within a report generation model~\cite{li2019knowledge, li2023auxiliary, zhang2020radiology, li2023dynamic, liu2021exploring}. Li~\textit{et al.}~\cite{li2019knowledge} compiled abnormalities from the MIMIC-CXR dataset, utilizing them as nodes in the abnormality graph they proposed. Zhang ~\textit{et al.}~\cite{zhang2020radiology} adopted a universal graph featuring 20 entities, connecting entities associated with the same organs. These relationships were incorporated into an adjacency matrix, utilized for message propagation in a graph encoding module. In contrast, Li~\textit{et al.}~\cite{li2023dynamic} dynamically updated the graph by injecting new knowledge extracted from the retrieved reports for each case.


\subsection{Retrieval-based Medical Report Generation}
To address the challenge of training a robust sentence generator, some studies~\cite{endo2021retrieval, han2018towards, ni2020learning, yang2021writing, transq} have turned to the retrieval of a database for generating descriptive sentences. This retrieval-based approach tends to better emulate the reporting practices of physicians and circumvent potential issues of model over-generation. Han~\textit{et al.}~\cite{han2018towards} introduced a symbol synthesis module that leveraged prior knowledge to transform the intention vector of observations into a dictionary. This dictionary, in combination with pre-defined rules, was used to revise the template. Endo~\textit{et al.}~\cite{endo2021retrieval} employed image-report representations pairs obtained from CLIP~\cite{radford2021learning} to retrieve unstructured, free-text radiology reports. Kong~\textit{et al.}~\cite{transq} and Gao~\textit{et al.}~\cite{gao2023simulating} proposed semantic query decoders, utilizing queries associated with different semantics to retrieve distinct content. This approach achieved state-of-the-art (SOTA) results across various metrics. Establishing a robust feature space representation is the foundation of the retrieval-based methods. 
It is worth noting that all the aforementioned methods equally retrieve common sentences and rare sentences without overcoming the issue of rare sentences being overshadowed. This is the main focus of this paper.

\section{Proposed Method}
\label{sec:method}

Figure~\ref{Fig:framework} shows the schematic illustration of Teaser. First, a radiological image is input into the Vision Transformer (ViT) to extract dense visual features. 
Subsequently, these features are fed into Abstracotr, which removes redundant and irrelevant information, resulting in more compact and task-relevant condensed visual features (refer to Section~\ref{subsec:vis}).
Next, the Topicwise Separable Encoder takes in several learnable queries and utilizes the sparse visual features obtained from Abstractor to generate diverse topic embeddings. 
%
These embeddings are then utilized to identify the most suitable sentences from the gallery, which includes all sentences from the training set. This process aims to pinpoint the most appropriate matching sentences for report generation.
Finally, a complete report is generated by filtering sentences based on topics and removing duplicates. 
In order to mitigate the issue of overlooking rare information during the retrieval process due to the long-tail distribution of data, we classify the learnable queries into two categories: common and rare.
The common queries are trained to match common medical observations with high-frequency in a group of medical reports such as pneumothorax and consolidation, while the rare queries are responsible for capturing descriptions specific to low-frequency yet crucial findings such as supraspinatus and osteoarthritis (refer to Section~\ref{subsec: difSR}).
Additionally, to enable different queries to observer different regions of image, we propose the Topic Contrastive Loss which ensures that each query learns unique topic (refer to Section~\ref{subsec:loss}.

\subsection{Visual Feature Extraction by Abstractor}
\label{subsec:vis}
We denote $\mathbf{I} \in \mathbb{R}^{C \times H \times W}$ as the input radiological image, where $[H, W]$ are height and width of $\mathbf{I}$, respectively. 
%
Subsequently, $\mathbf{I}$ is reshaped into the sequence of flattened 2D patches $\mathbf{I}_p \in \mathbb{R}^{N \times (P^2 C)}$ and fed into the visual encoder to extract dense visual features, where $P$ is the patch size, $N=\frac{HW}{P^2}$ is the number of patches. We use ViT-based~\cite{vit} as the visual encoder. Consequently, the latent image embeddings can be obtained by:
\begin{equation}
    \label{eq:ve}
     \mathbf{V} = \mathtt{ViT}(\mathbf{I}_p, [\mathrm{CLS}]),
\end{equation}
where $[\mathrm{CLS}]$ servers as the global embedding of an image. The sequence $\mathbf{V} \in \mathbb{R}^{(N+1) \times D}$ is the output of ViT, where $D$ is the dimension of the hidden state.

In previous report generation methods~\cite{R2Gen,multicriteria2022TMI}, $\mathbf{V}$ was directly fed into the decoder to generate medical reports. However, $\mathbf{V}$ is a high-dimensional and dense visual feature that often contains redundant and irrelevant information. The presence of such redundancy can hinder the proper expression of important anatomical regions, leading to inaccurate report generation. 
%
Recent research~\cite{jaegle2021perceiver,alayrac2022flamingo} has indicated that reducing the dimensionality of dense visual features is beneficial for extracting more relevant visual semantics while also reducing the computational complexity of vision-text cross-attention.

Motivated by this observation, we adopt an \textbf{Abstractor} that produces a small fixed number of visual embeddings when given a large number of input visual features.
This module improves the efficiency of subsequent visual-text cross-attention computations and reduces the noise in the image embeddings. As shown in Figure~\ref{Fig:framework}, the Abstractor consists of $L_a$ Transformer decoder layers. The learnable predefined number of latent input visual queries $\mathbf{Q}_{v} \in \mathbb{R}^{K \times D}$ and dense feature $\mathbf{V}$ are fed into Abstractor to obtain semantically condensed visual features:
\begin{equation}
    \label{eq:ve}
     \mathbf{A} = \mathtt{Abstractor}(\mathbf{Q}_v; 
\mathbf{V}), 
\end{equation}
where $K$ represents the number of latent input queries, which is significantly smaller than $N$, $\mathbf{A} \in \mathbb{R}^{K\times D}$ is the compressed visual features. 

\begin{figure}[]
  \centering
  \subfigure[w/o Abstractor]{
    \includegraphics[width=0.2\textwidth]{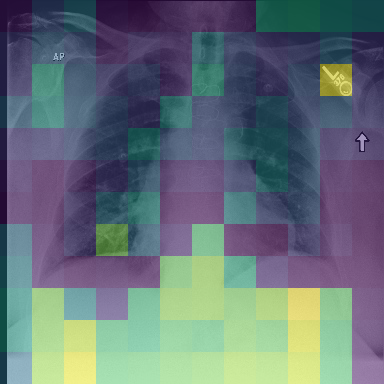}
    \label{fig:subfig1}
  }
  \subfigure[w/ Abstractor]{
    \includegraphics[width=0.2\textwidth]{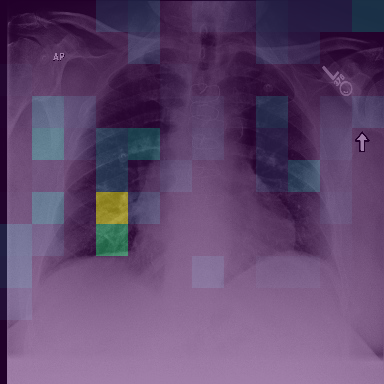}
    \label{fig:subfig2}
  }
  \caption{Attention maps with and without employing Abstractor.
  }
  \vspace{-5mm}
  \label{fig:abs}
\end{figure}

Figure~\ref{fig:abs} illustrates the attention maps of the model on various regions of the image with and without employing the Abstractor. It can be observed that without the Abstractor, the model exhibits relatively uniform attention across different regions of the image, even showing high attention scores on uninformative image edges. 
%
%
Conversely, when employing the Abstractor, a notable difference in attention is observed among regions, specifically with a high attention score on essential regions such as the lungs.
This highlights the crucial role of the Abstractor in extracting key visual information.



\subsection{Topicwise Separable Sentence Retrieval}
\label{subsec: difSR}
The previous retrieval-based report generation method~\cite{transq} utilizes a semantic encoder to capture the features of medical images. This is achieved by applying a set of semantic queries, where each query corresponds to a certain medical observation.
However, common topics (\textit{e.g.} pneumothorax and consolidation) that appear frequently in the training set are adequately trained, while rare topics (\textit{e.g.} supraspinatus and osteoarthritis) often receive insufficient attention and learning. 
The inherent long-tail distribution of medical data contribute to a phenomenon where the model tends to generate common findings while overlooking the rare ones, leading to inaccurate report generation.

To address this issue, we propose a \textbf{Topicwise Separable Encoder (TSE)} to achieve comprehensive report generation by separately querying the common and rare topics of the inputs. As illustrated in Figure~\ref{Fig:framework}, a set of learnable queries $\mathbf{Q}_c \in \mathbb{R}^{M \times D}$ representing common topics and another set $\mathbf{Q}_r \in \mathbb{R}^{J \times D}$ representing rare topics are fed into the Topicwise Separable Encoder composed of a $L$-layer transformer decoder block, where $M$, $J$ denote sequence length and $D$ denotes channel. Each block consists of a Multi-Head Cross-Attention (MCA) layer for interacting with visual features $\mathbf{A}$, a Multi-Head Self-Attention (MSA) for encoding the correlations between different topics, and a Multi-Layer Perceptron (MLP). We can obtain topic embeddings as follows:
\begin{equation}
\begin{aligned}
&\mathbf{H}_0 = \mathtt{MLP}([\mathbf{Q}_c, \mathbf{Q}_r]), \ \\
&\mathbf{H}_l^{'} = \mathtt{MSA}(\mathtt{LN}(\mathbf{H}_{l-1}))+\mathbf{H}_{l-1}, \quad l=1,...,L , \ \\
&\mathbf{H}_l^{''} = \mathtt{MCA}(\mathtt{LN}(\mathbf{H}_l^{'}), \mathtt{LN}(\mathbf{A})) + \mathbf{H}_l^{'}, \quad l=1,...,L , \ \\
&\mathbf{H}_l = \mathtt{MLP}(\mathtt{LN}(\mathbf{H}_l^{''})) + \mathbf{H}_l^{''}, \quad l=1,...,L, \ \\
&[\mathbf{T}_c, \mathbf{T}_r] = \mathtt{LN}(\mathbf{H}_L),
\end{aligned}
\end{equation}
where $\mathtt{LN}(\cdot)$ is layer normalization. $\mathbf{H}_l$, $\mathbf{H}_l^{'}$ and $\mathbf{H}_l^{''}$ represent middle-level features in $l$-th block. $\mathbf{T}_c \in \mathbb{R}^{M \times D}$ and $\mathbf{T}_r \in \mathbb{R}^{J \times D}$ denote the common and rare topic embeddings, respectively. To select the topics that are most relevant to the input image, we employ a linear projection to predict the selection probability $\mathbf{p} \in \mathbb{R}^{M+J}$ of candidate topics:
\begin{equation}
\mathbf{p} = \mathtt{MLP}([\mathbf{T}_c, \mathbf{T}_r]),
\end{equation}

\noindent\textbf{Criteria for discriminating common and rare sentences.} 
Sentences that fulfill both conditions are categorized as rare sentences: exhibiting significant semantic differences from other sentences and having lower occurrence frequencies. All remaining sentences are classified as common sentences.
Specifically, we discriminate these sentences by following process. 
First, we extract sentence embeddings from all the sentences in the training set using the pre-trained MPNet-v2~\cite{mpnet}. Next, we apply the K-Means algorithm~\cite{scikit-learn, kmeans2020} to cluster these sentence embeddings. By calculating the distances between each sentence embedding and the cluster centers, we identify the sentences that are distant from all the cluster centers (distances larger than 0.9 with each cluster center after normalization).  Moreover, we calculate the occurrence frequency for each sentence and identify those with a low frequency (frequency less than 6). 
Sentences that meet both these conditions are considered rare sentences; otherwise, are common sentences.

\noindent\textbf{In the training stage.} We aim to align the topic embeddings $\mathbf{T}_c$ and $\mathbf{T}_r$ with the sentence-level feature embeddings formed by the ground truth medical reports (denoted as $\mathbf{g}$). First, we partition $\mathbf{g}$ into common sentences $\mathbf{g}_c$ and rare sentences $\mathbf{g}_r$ based on aforementioned criteria. Subsequently, we utilize a pretrained SentenceEncoder to encode $\mathbf{g}_c$ and $\mathbf{g}_r$ into their respective sentence embeddings as follows:
\begin{equation} \label{equ:sentence_encoder}
[\mathbf{Y}_c, \mathbf{Y}_r] = \mathtt{SentenceEncoder}(\mathbf{g}_c, \mathbf{g}_r),
\end{equation}
where $\mathbf{Y}_c \in \mathbb{R}^{G_c \times D}$ and $\mathbf{Y}_r \in \mathbb{R}^{G_r \times D}$ are ground truth sentence embeddings, $G_c$ and $G_r$ denote the number of common and rare sentences, respectively. Here, we utilize MPNet-v2~\cite{mpnet} as the SentenceEncoder.

In general, the number of sentences in ground truth report is much smaller than the number of topic embeddings (\textit{i.e., $G_c < M$, $G_r < J$}). We employ the \textit{Hungarian matching}~\cite{kuhn1955hungarian} algorithm to select the embeddings from $\mathbf{T}_c$ and $\mathbf{T}_r$ that best match with $\mathbf{Y}_c$ and $\mathbf{Y}_r$, respectively:
\begin{equation}
    \begin{aligned}
        \hat{\sigma} = \begin{cases}
\underset{\sigma}{\mathtt{argmin}} \sum_{i=1}^{M} \mathcal{L}_\mathrm{e}(\mathbf{t}^{i}, \mathbf{y}^{\sigma (i)}), i \leq M  \\
\underset{\sigma }{\mathtt{argmin}} \sum_{i=M+1}^{M+J} \mathcal{L}_\mathrm{e}(\mathbf{t}^{i}, \mathbf{y}^{\sigma (i)}), i>M
\end{cases}
    \end{aligned}
\end{equation}
where $\sigma$ denotes all the matching strategies we are searching for, $\hat{\sigma}$ denote the optimal strategy that mapping of indices from topic embeddings $[\mathbf{t}^{1},...,\mathbf{t}^{M}] \in \mathbf{T}_c$, $[\mathbf{t}^{M+1},...,\mathbf{t}^{M+J}] \in \mathbf{T}_r$ to ground truth sentence embeddings $[\mathbf{y}^{\sigma(1)}, ..., \mathbf{y}^{\sigma(M)}] \in \mathbf{Y}_c$, $[\mathbf{y}^{\sigma(M+1)}, ..., \mathbf{y}^{\sigma(M+J)}] \in \mathbf{Y}_r$. 
$\mathcal{L}_\mathrm{e}$ represents the Euclidean distance metric.

By employing the matching strategy $\hat{\sigma}$, we obtain the optimal matching results $\mathbf{y}^{\hat{\sigma}(i)} $ for each topic embeddings $\mathbf{t}^{i}$. Then, the model is optimized using the matching results $\hat{\sigma}(i)$ by loss $\mathcal{L}$.
The overall loss $\mathcal{L}$ is composed of the topic alignment loss $\mathcal{L}_\mathrm{align}$ and the topic selection loss $\mathcal{L}_\mathrm{select}$.
\begin{equation}
    \mathcal{L} = \mathcal{L}_\mathrm{align} + \lambda\sum_{i=1}^{M+J}{\mathbbm{1}_{\hat\sigma(i) \neq \varnothing} \mathcal{L}_\mathrm{select}(\mathbf{c}^i, \mathbf{p}^{i})},
\end{equation}
where $\lambda$ is the trade-off parameter. $\mathbf{c}^i$ and $\mathbf{p}^i$ are the sentence selection label and selection probability in $i$-th index. For each $\mathbf{p}^i$, if its matching index $\hat\sigma(i) \neq \varnothing$, its corresponding sentence selection label $\mathbf{c}^i = 1$, and vice versa $\mathbf{c}^i = 0$. 
Considering the imbalanced distribution of topics, $\mathcal{L}_\mathrm{select}$ employs a multi-label classification loss called Distribution-Balanced Loss (DB-Loss)~\cite{wu2020distribution} to balance topic frequency differences.
The topic alignment loss $\mathcal{L}_\mathrm{align}$ will be introduced in Section~\ref{subsec:loss}.

\noindent\textbf{Report generation in the test stage.} We adhere to the standard retrieval approach, where we retrieve the most similar sentences from the encoded sentences within the training dataset for each topic embedding in $\mathbf{T}_c$ and $\mathbf{T}_r$. This process culminates in the creation of the final medical report. First, we form the sentence galleries, $\mathcal{D}_c$ and $\mathcal{D}_r$, by encoding all common sentences and rare sentences, respectively, from the training set using a pre-trained sentence encoder (similar to Equation~\ref{equ:sentence_encoder}). Next, we compute the similarity between each topic embedding $\mathbf{t}^i$ and the two galleries, selecting the sentences with the highest similarity scores.
\begin{equation}
\begin{aligned}
    \hat{j} &=& &\underset{j}{\mathtt{argmax}}({\mathbf{t}^i \cdot \mathbf{d}_c^{j}}), \mathbf{p}^i > \tau_c, i \leq M, \ \\
    \hat{k} &=& &\underset{k}{\mathtt{argmax}}({\mathbf{t}^i \cdot \mathbf{d}_r^{k}}), \mathbf{p}^i > \tau_r, i > M,
\end{aligned}
\end{equation}
where $\mathbf{d}_c^{j} \in \mathbb{R}^{D}$ and $\mathbf{d}_r^{k} \in \mathbb{R}^{D}$ denotes each sentence embedding in $\mathcal{D}_c$ and $\mathcal{D}_r$, respectively. $\tau_c$ and $\tau_r$ denote the probability thresholds for selecting candidate common and rare topics, respectively. Finally, these retrieved sentences are combined and sorted~\cite{transq} to form the final report.

\subsection{Topic Alignment Loss}
\label{subsec:loss}

The previous work~\cite{transq} employs the cosine similarity loss to achieve alignment loss $\mathcal{L}_{\mathrm{align}}$ between the topic embeddings $\mathbf{t}^{i}$ and the matching ground truth sentence embeddings $\mathbf{y}^{\hat{\sigma}(i)}$. The similarity loss is used to measure the similarity between two embeddings by calculating the cosine of the angle between them as following:
\begin{equation}
\label{eq:sim}
\mathcal{L}_\mathrm{sim}=\sum_{i=1}^{M+J}\left ( 1-\frac{\mathbf{t}^{i}\cdot \mathbf{y}^{\hat{\sigma}(i)} }{\left \| \mathbf{t}^{i} \right \|\cdot \left \| \mathbf{y}^{\hat{\sigma}(i)} \right \|  } \right ),
\end{equation}
where, $[\mathbf{t}^{1},...,\mathbf{t}^{M}] \in \mathbf{T}_r$, $[\mathbf{t}^{M+1},...,\mathbf{t}^{M+J}] \in \mathbf{T}_s$.



\begin{figure}[]
  \centering
  \subfigure[w/o TCL]{
    \includegraphics[width=0.46\linewidth]{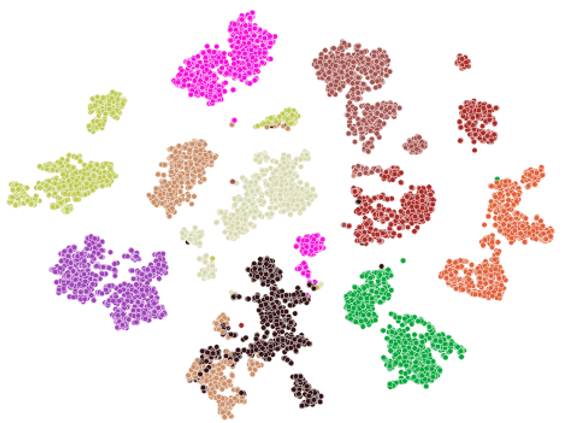}
    \label{fig:subfig1}
  }
  \subfigure[w/ TCL]{
    \includegraphics[width=0.46\linewidth]{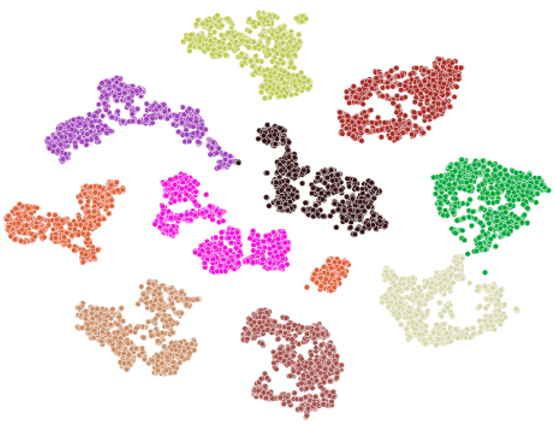}
    \label{fig:subfig2}
  }
  \caption{t-SNE of topic embeddings with and without TCL. The topic embeddings generated by the identical query are represented using the identical color.}
  \vspace{-5mm}
  \label{fig:tsne}
\end{figure}

However, the aforementioned alignment loss only focuses on pulling the ground truth and positive samples (matched topic embeddings) together, without pushing negative samples (unmatched topic embeddings) far away. This may result in learning an ambiguous latent space, thereby leading to issues such as many-to-one or one-to-many mappings.
%
Figure~\ref{fig:tsne}~(a) illustrates the distribution of topic embeddings corresponding to different queries in the feature space. It is evident that certain topic embeddings generated from distinct queries exhibit substantial overlap in the latent space. Consequently, this overlap leads to the learning of the same sentence being associated with different topics.

To mitigate this issue, we introduce the \textbf{Topic Contrastive Loss (TCL)}, which aims to bring embeddings of the same topic into closer proximity in the latent space while simultaneously pushing away embeddings from different topics. This ensures that each query is dedicated to expressing its respective specialized topic.
Specifically, for each ground truth sentence embedding $\mathbf{y}^{\hat{\sigma}(i)}$ in the report, we consider its matching topic embedding $\mathbf{t}^{i}$ as a positive example, and the other sentences in the report as negative examples, we obtain topic embeddings to ground truth sentence embeddings $\underset{t\to r}{\mathcal{L}_{\mathrm{TCL} } }$ as:
\begin{equation}
    \underset{t\to r}{\mathcal{L}_{\mathrm{TCL} } } =\sum_{i=1}^{M+J} \left ( -log\frac{exp(\mathbf{t}^{i}\cdot \mathbf{y}^{\hat{\sigma}(i)})/\tau }{\sum_{j=1}^{M+J}(exp(\mathbf{t}^{i}\cdot \mathbf{y}^{\hat{\sigma}(j)}/\tau )} \right ),
    \label{eq:TCL}
\end{equation}
where $\tau$ denotes the temperature coefficient. 
Similarly, we can obtain the TCL from ground truth sentence embeddings to topic embeddings $\underset{r \to t}{\mathcal{L} _{\mathrm{TCL} }}$. Thus, the overall topic contrastive loss $\mathcal{L} _{\mathrm{TCL} }$ can be formulated as follow:
\begin{equation}
    \label{eq:TCL}
    \mathcal{L} _{\mathrm{TCL} }=\underset{t \to r}{\mathcal{L} _{\mathrm{TCL} }} + \underset{r \to t}{\mathcal{L} _{\mathrm{TCL} }} .
\end{equation}
By introducing TCL, we observe an improvement in the aggregation of topic embeddings (as shown in Figure~\ref{fig:tsne}~(b)), leading to a mitigation of the confounding issue.

The total alignment loss is formulated as:
\begin{equation}
    \mathcal{L}_{\mathrm{align}} =\mathcal{L}_{\mathrm{sim}}+\alpha \mathcal{L} _{\mathrm{TCL} },
\end{equation}
where $\alpha$ is the trade-off parameter.

\section{Experiments}
\label{sec:experiments}
\subsection{Datasets}
\label{subsec:dataset}

We evaluate the effectiveness of our proposed Teaser on two public radiology report generation benchmarks, i.e., \textbf{MIMIC-CXR}~\cite{johnson2019mimic} and \textbf{IU-Xray}~\cite{iuxraydataset}. 

\subsubsection{MIMIC-CXR}
MIMIC-CXR dataset~\cite{johnson2019mimic} is the largest public radiology dataset, consisting of 377,110 chest X-ray images and 227,835 radiology reports from 64,588 patients.
%
Images are provided with 14 labels derived from NegBio~\cite{peng2018negbio} and CheXpert~\cite{irvin2019chexpert} applied to the corresponding free-text radiology reports.
In our experiment, we follow the official split to ensure a fair comparison, dividing the dataset into 222,758 training samples, 1,808 validation samples, and 3,269 testing samples.
This adherence to the official split allows for consistent evaluation and comparison of results with other studies using the same dataset.

\subsubsection{IU-Xray}
IU-Xray dataset~\cite{iuxraydataset} has been extensively employed for evaluating the performance of radiology reporting methods. This dataset consists of 3,955 radiology reports and 7,470 chest X-ray images. Each report is associated with either frontal or frontal and lateral view images. 
To ensure the fairness and credibility of the experimental results, we adopt the commonly used split methods~\cite{R2Gen,liu2021exploring} to divide the dataset into training (70\%), validation (10\%), and testing (20\%) sets. This approach is widely accepted in the research community for evaluating the performance of various models.  

\subsection{Evaluation Metrics}
\label{subsec:metrics}
To evaluate the performance of our proposed method, we employ the widely-used natural language generation (NLG) metrics and clinical efficacy (CE) metrics.
For NLG metrics, we adopt the standard evaluation protocol to evaluate the generated diagnostic reports: BLEU~\cite{bleu}, METEOR~\cite{banerjee2005meteor}, ROUGE-L~\cite{lin2004rouge}.
For the description of clinical abnormalities in the generated reports, we compare the generated reports with the pseudo-labels generated by CheXpert~\cite{irvin2019chexpert} in 14 different categories. To evaluate the performance of the model in terms of clinical efficacy, we employ precision, recall, and F1 scores~\cite{yan2023attributed}. These metrics provide a comprehensive assessment of how well the model captures and describes the clinical abnormalities in the reports, enabling a robust evaluation of its performance.

\subsection{Implementation details} 
\label{subsec:implementation}
We utilize the tokenizer from Hugging Face's transformer library~\cite{huggingface}, specifically the ``bert-base-uncased'' model, to tokenize all words in the reports for both datasets.
To obtain sentence embeddings, we employ a pre-trained MPNet-v2 model~\cite{mpnet}.
As the visual encoder, we employ a vision transformer model i.e., ViT-B/32, which has been pre-trained on the Image-Net dataset. The input image size [C, H, W] is set to 3$\times$384$\times$384, the patch size $P$ is 32, and the hidden channel size $D$ is 768. The number of Transformer decoder layers $L_a$ and $L$ in Abstractor and TSE are both set to 12. The number of visual queries $K$ is set to 64.
We utilize paired images of a patient as the input for IU X-ray and one image for MIMIC-CXR. 
For the MIMIC-CXR dataset, we set the number of common and rare topic queries to $M=50$ and $J=20$, respectively. The sentence selection thresholds are set to $\tau_c=\tau_r=0.38$.
For the IU X-ray dataset, we set them to $M=25$, $J=8$, and $\tau_c=\tau_r=0.45$.
In both datasets, we set the scale factor of loss function $\lambda=1$ and $\alpha=0.1$.
Our model training process utilizes the AdamW optimizer~\cite{adam} with a learning rate of 1e–4 for MIMIC-CXR, 1e-5 for IU-Xray, applying linear decay for parameter optimization. The batch size is set to 64 for the MIMIC-CXR dataset and 32 for the IU X-ray dataset. We implement our model using Pytorch and Pytorch-lightning on NVIDIA GeForce RTX 3090 GPU.

\subsection{Quantitative Comparison with SOTA Methods}

\begin{table*}[!t]
    \centering
    \caption{        
    Comparison on~\textbf{MIMIC-CXR} dataset. The best values are highlighted in bold.}

    \label{tab:mimic_sota}
    \renewcommand\arraystretch{1.1}
    \resizebox{0.96\textwidth}{!}{
        \begin{tabular}{c|c c c|c c c c c c}
            \hline
            \hline
            \multirow{2}{*}{Method} & \multicolumn{3}{c|}{CE Metrics} & \multicolumn{6}{c}{NLG Metrics} \\
            & Precision & Recall & F1 & BLEU-1 & BLEU-2 & BLEU-3 & BLEU-4 & METEOR & ROUGE \\
            \hline
            R2Gen~\cite{R2Gen} & 0.333 & 0.273 & 0.276 & 0.353 & 0.218 & 0.145 & 0.103 & 0.142 & 0.277 \\
            PPKED~\cite{liu2021exploring} & - & - & - & 0.360 & 0.224 & 0.149 & 0.106 & 0.149 & 0.284 \\
            CA~\cite{liu2021CA} & 0.352 & 0.298 & 0.303 & 0.350 & 0.219 & 0.152 & 0.109 & 0.151 & 0.283 \\
            KGAE~\cite{KGAE} & 0.389 & 0.362 & 0.355 & 0.369 & 0.231 & 0.156 & 0.118 & 0.153 & \textbf{0.295} \\
            Multicriteria~\cite{multicriteria2022TMI} & - & - & - & 0.351 & 0.223 & 0.157 & 0.118 & - & 0.287 \\
            KnowMT~\cite{KnowMT2022MIA} & 0.458 & 0.348 & 0.371 & 0.363 & 0.228 & 0.156 & 0.115 & - & 0.284 \\
            TranSQ~\cite{transq} & 0.420 & 0.500 & 0.457 & 0.416 & 0.252 & 0.161 & 0.107 & 0.169 & 0.262 \\
            ICT~\cite{ICT2023JBHI} & - & - & - & 0.376 & 0.233 & 0.157 & 0.113 & 0.144 & 0.276 \\
            DCL~\cite{DCL2023CVPR} & 0.471 & 0.352 & 0.373 & - & - & - & 0.109 & 0.150 & 0.284 \\
            METransformer~\cite{METransformer2023CVPR} & - & - & - & 0.386 & 0.250 & \textbf{0.169} & \textbf{0.124} & 0.152 & 0.291 \\
            RAMT~\cite{RAMT2023TMM} & 0.380 & 0.342 & 0.335 & 0.362 & 0.229 & 0.157 & 0.113 & 0.153 & 0.284 \\
            MedXChat~\cite{MedXChat2023arXiv} & - & - & - & 0.367 & 0.235 & 0.158 & 0.111 & 0.135 & 0.264 \\
            RaDialog~\cite{RaDialog2023arXiv} & - & - & 0.394 & 0.346 & - & - & 0.095 & 0.140 & 0.271 \\
            PromptMRG~\cite{Promptmrg} & 0.501 & 0.509 & 0.476 & 0.398 & - & - & 0.112 & 0.157 & 0.268 \\
            \hline
            Teaser (Ours) & \textbf{0.534} & \textbf{0.518} & \textbf{0.526} & \textbf{0.423} & \textbf{0.257} & 0.166 & 0.113 & \textbf{0.170} & 0.287 \\
            \hline
            \hline
        \end{tabular}
    }
\end{table*}
\begin{table}[!t]
    \centering
    \caption{
        Comparison on~\textbf{IU-Xray} dataset. The best values are highlighted in bold.}
    \label{tab:iuxray_sota}
    \renewcommand\arraystretch{1.2}
    \resizebox{0.5\textwidth}{!}{
        \begin{tabular}{c|c c c c c c}
            \hline
            \hline
             Method & BLEU-1 & BLEU-2 & BLEU-3 & BLEU-4 & METEOR & ROUGE \\
            \hline
            R2Gen~\cite{R2Gen} & 0.470 & 0.304 & 0.219 & 0.165 & 0.187 & 0.371 \\
            PPKED~\cite{liu2021exploring} & 0.483 & 0.315 & 0.224 & 0.168 & 0.190 & 0.376 \\
            CA~\cite{liu2021CA} & 0.492 & 0.314 & 0.222 & 0.169 & 0.193 & 0.381\\
            KGAE~\cite{KGAE} & 0.519 & 0.331 & 0.235 & 0.174 & 0.191 & 0.376\\
            Multicriteria~\cite{multicriteria2022TMI} & 0.496 & 0.319& 0.241 & 0.175 & - & 0.377 \\
            KnowMT~\cite{KnowMT2022MIA} & 0.496 & 0.327 & 0.238 & 0.178 & - & 0.381 \\
            TranSQ~\cite{transq} & 0.484 & 0.333 & 0.238 & 0.175 & 0.207 & 0.415 \\
            ICT~\cite{ICT2023JBHI} & 0.503 & 0.341 & 0.246 & 0.186 & 0.208 & 0.390 \\
            DCL~\cite{DCL2023CVPR} & - & - & - & 0.163 & 0.193 & 0.383 \\
            METransformer~\cite{METransformer2023CVPR} & 0.483 & 0.322 & 0.228 & 0.172 & 0.192 & 0.380 \\
            RAMT~\cite{RAMT2023TMM} & 0.482 & 0.310 & 0.221 & 0.165 & 0.195 & 0.377 \\
            PromptMRG~\cite{Promptmrg} & 0.401 & - & - & 0.098 & 0.160 & 0.281 \\
            \hline
            Teaser (Ours) & \textbf{0.527} & \textbf{0.360} & \textbf{0.260} & \textbf{0.192} & \textbf{0.211} & \textbf{0.417} \\
            \hline
            \hline
        \end{tabular}
    }
\end{table}

To demonstrate the effectiveness of our model, 
we conduct a comprehensive comparison with a wide range of SOTA report generation models on both two datasets,
%
including:
(1) LSTM-based model like CA~\cite{liu2021CA},
(2) generation-based models, i.e. R2Gen~\cite{R2Gen}, PPKED~\cite{liu2021exploring}, Multicriteria~\cite{multicriteria2022TMI}, KnowMT~\cite{KnowMT2022MIA}, ICT~\cite{ICT2023JBHI}, METransformer~\cite{METransformer2023CVPR}), PromptMRG~\cite{Promptmrg}
(3) knowledge-graph-enhanced models such as KGAE~\cite{KGAE}, DCL~\cite{DCL2023CVPR}, RAMT~\cite{RAMT2023TMM},
(4) retrieval-based model like TranSQ~\cite{transq},
(5) LLM-based models, i.e. MedXChat~\cite{MedXChat2023arXiv}, RaDialog~\cite{RaDialog2023arXiv}.
By observing results of Teaser compared with SOTA, we have observed the following findings.

Firstly, on the MIMIC-CXR dataset (refer to Table~\ref{tab:mimic_sota}), our method outperforms all the compared SOTA methods in CE metrics.
Specifically, our Precision metric increases from 0.501 to 0.534, Recall increases from 0.509 to 0.518, and F1-score increases from 0.476 to 0.526. 
%
The substantial increase in Precision suggests that our model has a lower false positive rate for the 14 disease categories.
The increase in Recall indicates that the model can correctly detect true positive cases while reducing the number of missed positive cases. The high F1-score, which considers both Precision and Recall, demonstrates the model's good performance in balancing the prediction of Precision and Recall.
Above performance indicates that our model, incorporating topicwise separable sentence retrieval strategy can allocate more attention to rare topics, which achieves more accurate detection of findings.

Secondly, our method not only exhibits favorable performance in the CE metrics but also demonstrates outstanding results across various NLG metrics. Table~\ref{tab:iuxray_sota} reveals that our Teaser surpasses SOTA approaches in all NLG metrics on the IU-Xray dataset, with improvements ranging from 1.5\% to 5.7\%.
%
%
Furthermore, on the MIMIC-CXR dataset, our method achieves the highest scores in BLEU-1 (0.423), BLEU-2 (0.257), and METEOR (0.170) metrics. These findings suggest that the reports generated by our Teaser align well with the vocabulary, semantics, and grammar utilized by medical professionals. Teaser exhibits exceptional performance in generating natural language outputs, which can be attributed to its sentence retrieval-based algorithm.

\subsection{Visual Comparison with SOTA Methods}
\label{subsec:visualCom}

\begin{figure}[!t]
\centering
\includegraphics[width =1\linewidth]{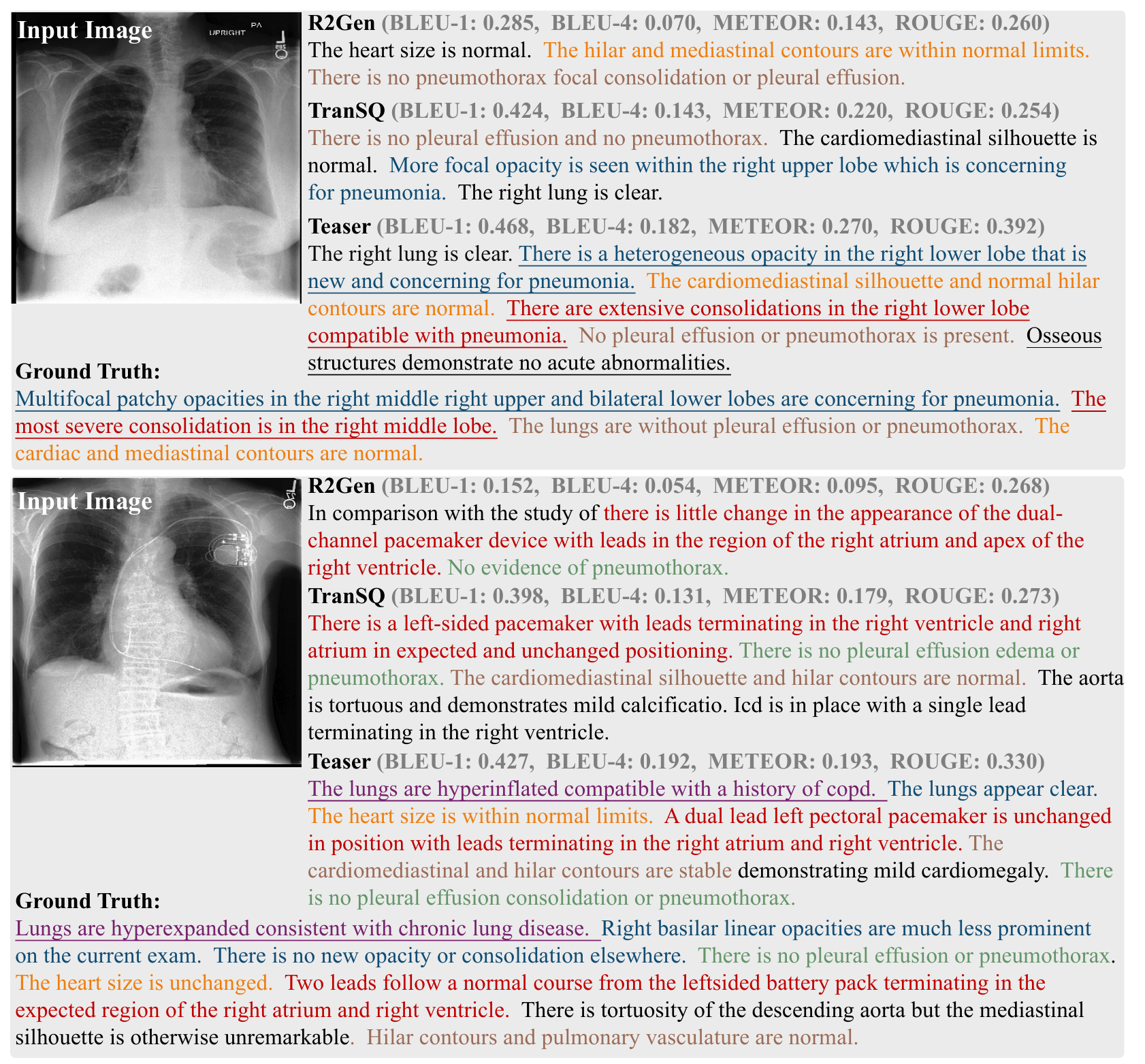}
\caption{Visualization and reports comparison between our Teaser and SOTA Methods. Sentences shaded in the same color represent corresponding to the same topics. The underlined texts indicate the rare sentences.}
\vspace{-5mm}
\label{Fig:visualisationSOTA}
\end{figure}

Figure~\ref{Fig:visualisationSOTA} shows the results of our qualitative comparison by comparing Teaser with R2Gen~\cite{R2Gen}, TranSQ~\cite{transq} on the MIMIC-CXR dataset. 
In the first case, R2Gen fails to describe the existing abnormal findings in the image (in blue), while TranSQ mentions the focal opacity in the right upper lobe but lacks a description of the consolidation (in red). In contrast, our Teaser generates descriptions that accurately align with the medical findings in the image, covering both aspects of information.
Similarly, in the second case, neither R2Gen nor TranSQ provides descriptions of rare parts (in purple). However, our method accurately outputs the rare sentence \textit{``The lungs are hyperinflated compatible with a history of COPD."}
These findings demonstrate the excellent performance of our proposed ToDi method in describing rare information.

\subsection{Ablation Analysis}
\label{sec:ablation}
\begin{table*}[!t]
    \centering
    \caption{
        Ablation studies on MIMIC-CXR. The ``BASE'' represents model only composed by ViT and semantic encoder. Here, ``TSE'', ``TCL'', and ``ABS''stand for ``'Topicwise Separable Encoder', ``Topic Contrastive Loss'', and ``Abstractor'', respectively.}
    \label{tab:mimic_ablation}
    \renewcommand\arraystretch{1.2}
    \resizebox{0.9\textwidth}{!}{
        \begin{tabular}{c l | c c c| c c c c c c}
            \hline
            \hline
            \multirow{2}{*}{\#} & \multirow{2}{*}{Models} & \multicolumn{3}{c|}{CE Metrics} & \multicolumn{6}{c}{NLG Metrics} \\
            & & Precision & Recall & F1 & BLEU-1 & BLEU-2 & BLEU-3 & BLEU-4 & METEOR & ROUGE \\
            \hline
            1 & BASE & 0.415 & 0.431 & 0.423 & 0.407 & 0.236 & 0.145 & 0.091 & 0.158 & 0.255 \\

            2 & BASE+TSE & 0.468 & 0.512 & 0.489 & 0.408 & 0.247 & 0.158 & 0.105 & 0.163 & 0.260 \\

            3 & BASE+TSE+TCL & 0.516 & 0.511 & 0.513 & 0.419 & 0.251 & 0.160 & 0.107 & 0.168 & 0.260\\
            4 & BASE+TSE+TCL+ABS & \multirow{2}{*}{\textbf{0.534}} & \multirow{2}{*}{\textbf{0.519}} & \multirow{2}{*}{\textbf{0.526}} & 0.423 & 0.255 & 0.163 & 0.110 & 0.169 & 0.258\\
            
            5 & BASE+TSE+TCL+ABS+SORT & & & &  \textbf{0.423} & \textbf{0.257} & \textbf{0.166} & \textbf{0.113} & \textbf{0.170} & \textbf{0.287} \\				
            \hline
            \hline
        \end{tabular}
    }
\end{table*}

\begin{figure}
    \centering
    \includegraphics[width =1\linewidth]{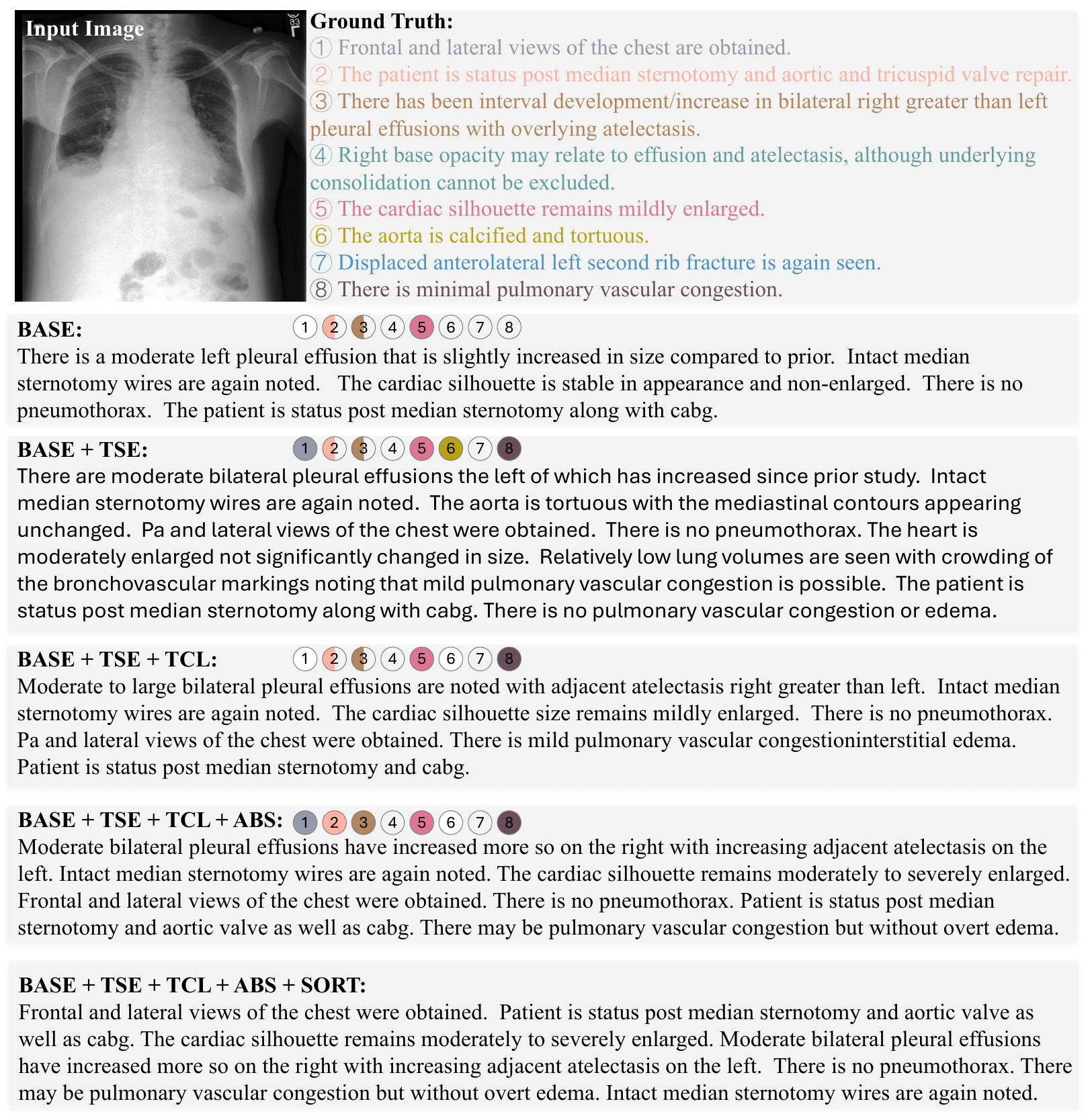}
    \caption{Visualization of the Ablation Study. The numbers in models correspond to the indices of the ground truth sentences. The degree of coloring is directly proportional to the semantic consistency between the generated sentences and the ground truth. These consist of semantic consistency, partial consistency, and semantic inconsistency, respectively.}
    \label{fig:ablation}
    \vspace{-2mm}
\end{figure}

On MIMIC-CXR, we conduct ablative experiments on the key components of our proposed model across various settings. The quantitative comparison is presented in Table~\ref{tab:mimic_ablation}, while qualitative comparison is shown in Figure~\ref{fig:ablation}.
%
%
%
First, we consider the model with the visual and semantic encoders as the baseline, named as \textbf{BASE}.
Second, we introduce the proposed Topicwise Separable Encoder (TSE) in Section~\ref{subsec: difSR} as \textbf{BASE+TSE} in the second row of Table~\ref{tab:mimic_ablation}. BASE+TSE achieves a significant improvement in CE metrics and provides more comprehensive descriptions. 
Specifically, BASE does not provide descriptions for uncommon information such as ``\textit{the aorta is calcified and tortuous}" or ``\textit{pulmonary vascular congestion}". By introducing TSE, the aforementioned issue has been solved. However, the generated reports exhibit semantic repetition, such as the description of ``\textit{intact median sternotomy wires}". There are also contradictory statements between ``\textit{mild pulmonary vascular congestion is possible}" and ``\textit{There is no pulmonary vascular congestion or edema}".
Building upon BASE+TSE, we introduce the Topic Contrastive Loss (TCL), namely \textbf{BASE+TSE+TCL} (i.e. Row 3). In this setting, the BLEU-1 metric exhibits the most significant improvement, rising from 0.408 to 0.419. Instances of repetition and contradictory descriptions in BASE+TSE are eliminated. 
This is attributed to TCL in bringing similar topic embeddings together and pushing them away from embeddings of other topics. 
However, in BASE+TSE+TCL, only the information regarding ``\textit{post median sternotom}" is described, while ``\textit{aortic and tricuspid valve repair}" is missing.
Subsequently, in \textbf{BASE+TSE+TCL+ABS} (i.e. Row 4), by introducing Abstractor (ABS), the model focus more on the relevant regions, leading to a notable improvement in CE and NLG metrics.
%
Finally, we performed a reordering of the generated sentences (refer to Row 5), resulting in a sentence sequence that better aligns with the writing conventions of medical professionals.

\section{Discussion}
\label{sec:discussion}
\subsection{Image-Sentence Alignment of Teaser}
\label{sec:visualisation}

\begin{figure*}[!t]
\centering
\includegraphics[width =1\linewidth]{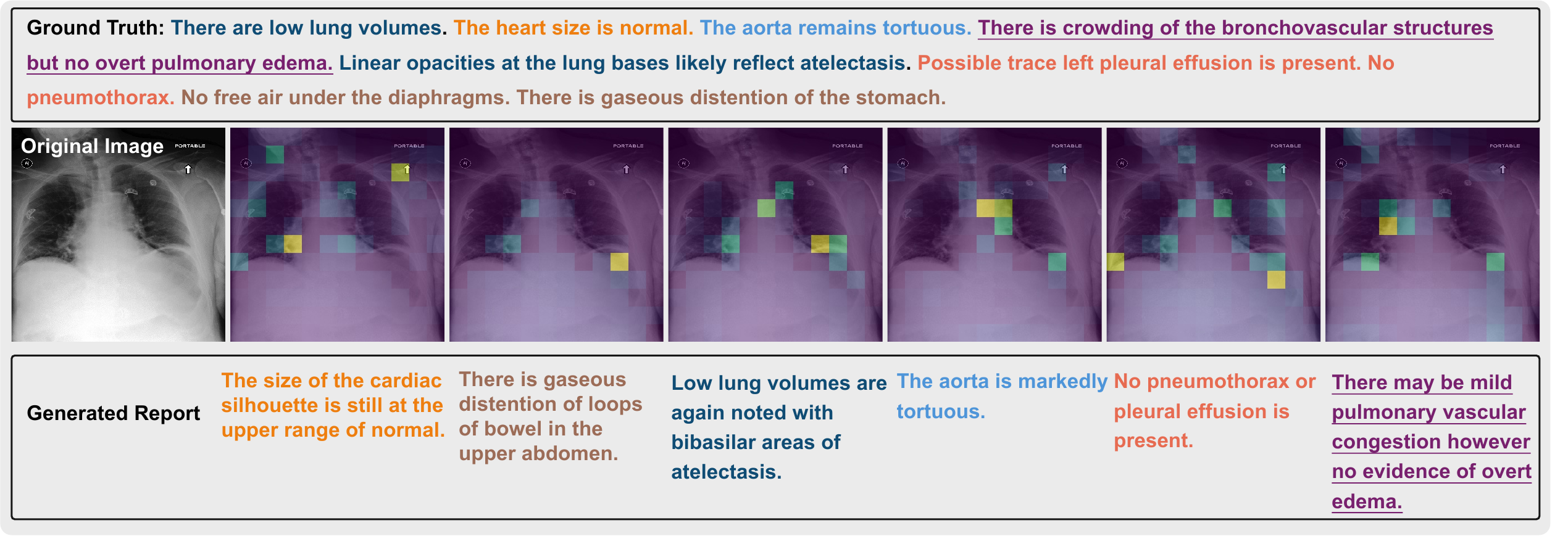}
\caption{
Alignment between sentences and their corresponding attention maps.
Different colors are used to highlight the pairwise ground truth sentence and generated sentence, indicating their topic alignment. The rare sentences are underlined.}
\vspace{-5mm}
\label{Fig:visualisation}
\end{figure*}

To further demonstrate the performance of the proposed Teaser, we analyze the visual case from the MIMIC-CXR dataset in Figure~\ref{Fig:visualisation}. 
%
%
Firstly, we observe a strong topic alignment between the generated sentences and ground truth report.
Secondly, the generated report and the attention map demonstrate a strong correspondence, which indicates that Teaser effectively captures visual features and achieves alignment between visual embeddings and topic embeddings.
Thirdly, our model is capable of generating rare sentences which are underlined in this Figure. This indicates that our model effectively captures and generates rare information with low frequency in the medical report dataset, maintaining the overall semantic coherence.

\subsection{The Number of Visual Queries in Abstractor}
\label{sec:abstractor}

\begin{table}[!t]
    \centering
    \caption{Performance comparison on MIMIC-CXR with different number $K$ of visual queries in Abstractor.}
    \label{tab:abstractorTokens}
    \renewcommand\arraystretch{1.2}
    \resizebox{0.48\textwidth}{!}{
        \begin{tabular}{c|c c c|c c c}
            \hline
            \hline
             \multirow{2}{*}{$K$} & \multicolumn{3}{c|}{CE Metrics} & \multicolumn{3}{c}{NLG Metrics} \\
              & Precision & Recall & F1 & BLEU-1 & BLEU-4 & ROUGE\\
             \hline
             0 & 0.408 & 0.431 & 0.419 & 0.368 & 0.070 & 0.233 \\
             32 & 0.522 & 0.506 & 0.514 & 0.422 & 0.108 & \textbf{0.259}\\
             \textbf{64 (ours)} & \textbf{0.534} & 0.519 & \textbf{0.526} & \textbf{0.423} & \textbf{0.110} & 0.258\\
             128 & 0.527 & \textbf{0.520} & 0.523 & 0.419 & 0.108 & 0.258\\
            \hline
            \hline
        \end{tabular}
    }
    \vspace{-2mm}
\end{table}


The number of visual queries $K$ in Abstractor affects its ability in visual feature extraction and abstraction, as well as the dimensionality of the compressed visual features \textbf{A}. It also influences the feature encoding and computational complexity of the subsequent topic encoder, which has a certain impact on the performance of Teaser.
Specifically, when the value of $K$ is small, the obtained compressed features \textbf{A} will remove more redundant information, thereby reducing the difficulty of retrieving corresponding sentences for the subsequent topic encoder. 
Nevertheless, if $K$ is excessively decreased, it may lead to insufficient visual features to support subsequent report generation.
If $K$ is excessively increased, the obtained visual features will have higher granularity, but it also implies higher feature dimensionality, which increases the difficulty of subsequent sentence retrieval.

\begin{figure}[]
    \centering
    \includegraphics[width =1\linewidth]{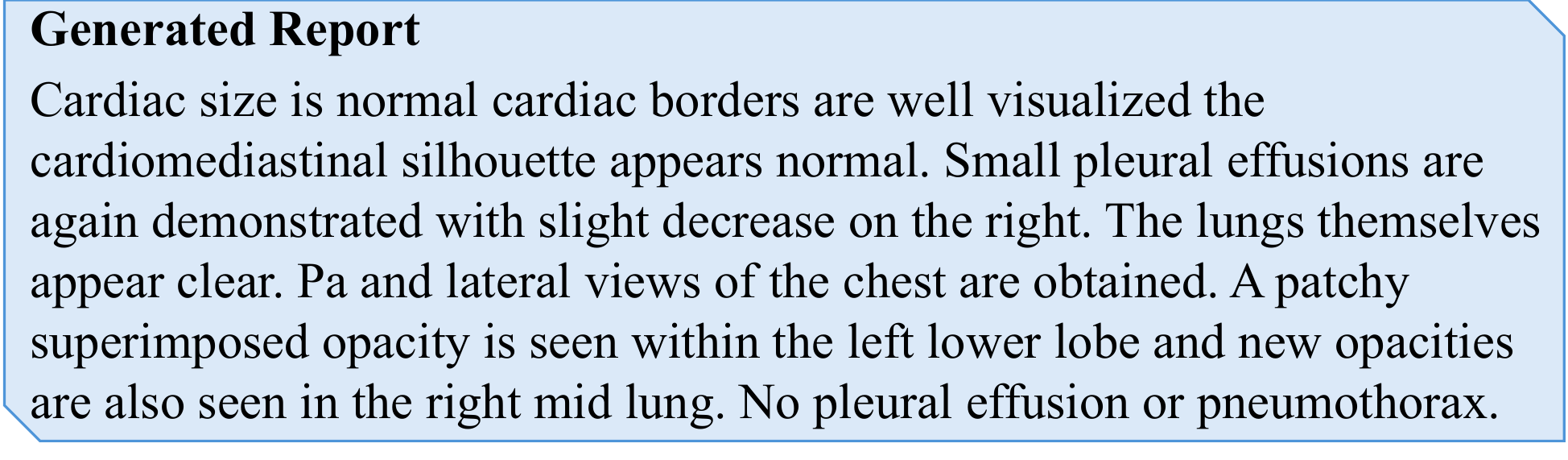}
    \vspace{-5mm}
    \caption{
    %
    Generated report without visual embeddings.
    }
    \label{fig:report_template}
\end{figure}

In this section, we keep other parameter settings and discuss the performance of Teaser at $K=\{0, 32, 64, 128\}$.
It is important to note that $K=0$ means the learnable visual queries $\mathbf{Q}_{v}$ in Abstractor are empty. 
Consequently, for any given input image, the Topic Separable Encoder generates the medical report without visual feature input. This indicates that the generated reports heavily depend on the topic features learned by the topic encoder, which can be considered as template features.
The generated template report is shown in Figure~\ref{fig:report_template}. F1 score of 0.419 and the BLEU-1 score of 0.368 represent the lower bounds of performance for retrieval-based methods on this dataset (refer to Table~\ref{tab:abstractorTokens}).

The experimental results for $K=\{0, 32, 64, 128\}$ are presented in Table~\ref{tab:abstractorTokens}, and the best performance is observed at $K = 64$ across almost all metrics. Consequently, we adopt $K=64$ in our Teaser architecture.

\subsection{Sensitivity to the Number of Rare Topic Queries}
\label{sec:specific}

\begin{table}[!t]
    \centering
    \caption{Generation performance on MIMIC-CXR with different number $J$ of rare queries.}    \label{tab:specificCount}
    \renewcommand\arraystretch{1.2}
    \resizebox{0.48\textwidth}{!}{
        \begin{tabular}{c|c c c|c c c}
            \hline
            \hline
             \multirow{2}{*}{$J$}& \multicolumn{3}{c|}{CE metrics} & \multicolumn{3}{c}{NLG metrics} \\
             & Precision & Recall & F1 & BLEU-1 & BLEU-4 & ROUGE\\
             \hline
             0 & 0.524 & 0.492 & 0.508 & 0.406 & 0.094 & 0.250\\
		
             10 & 0.515 & 0.501 & 0.507 & 0.419 & 0.106 & \textbf{0.260}\\
             
            \textbf{20 (ours)} & \textbf{0.534} & \textbf{0.519} & \textbf{0.526} & \textbf{0.423} & \textbf{0.110} & 0.258\\

             30 & 0.524 & 0.495 & 0.509 & 0.415 & 0.108 & 0.259\\

            \hline
            \hline
        \end{tabular}
    }
    \vspace{-4mm}
\end{table}

In this section, we conduct experiments using $J=\{0,10,20,30\}$ to evaluate how the number of rare topic queries affects the performance of Teaser in Table~\ref{tab:specificCount}. To ensure accurate comparison, we keep consistency in other experimental parameters.
$J = 0$ means all rare sentences in the training set are removed, only common sentences are retained. 
Compared to other settings, $J=0$ exhibits poor results in these metrics.
It indicates that rare queries often retrieval indicate critical findings that should be mentioned in the generated report.
As the $J$ increases, the Teaser shows an overall trend of initially increasing and then decreasing in terms of the CE and NLG metrics. Additionally, the overall performance is optimal when $J=20$. We speculate that this might be due to the increased semantic diversity and redundancy in the model's expressions as $J$ grows, thereby providing richer details. However, this also increases the possibility of different topic queries retrieving the same topics. Consequently, the selection of subsequent sentences becomes more challenging, leading to a higher risk of making incorrect choices. 
Similarly, when $J$ decreases, there is a possibility of incomplete semantic coverage for specific topic queries, resulting in a decrease in the diversity of generated reports.
Taking into account the experimental results and our speculative analysis, we ultimately set $J=20$ in our final model.

\section{Conclusion}
\label{sec:conclusion}
This paper proposes a topicwise separable sentence retrieval method for medical report generation. By adopting a strategy that separates common topic queries from rare topic queries, we successfully address the issue of conventional retrieval-based methods being unable to learn rare topics. Our proposed TCL effectively aligns similar semantic queries and topics, reducing the confusion in matching between queries and topics. Furthermore, the introduced Abstractor module reduces visual noise, thereby aiding the topic decoder in better understanding the visual observational intents. Our proposed method can generate more accurate and comprehensive medical reports and achieves state-of-the-art results on two publicly available datasets. For example, on the MIMIC-CXR dataset, it exhibits a significant improvement of 11\% in terms of the F1 score.

\bibliographystyle{IEEEtran}
\bibliography{RG}

\end{document}